\documentclass[runningheads]{llncs}

\usepackage{eccv}

\usepackage{eccvabbrv}

\usepackage{graphicx}
\usepackage{booktabs}

\usepackage{multirow}
\usepackage{colortbl}

\usepackage[accsupp]{axessibility}  

\usepackage[pagebackref,breaklinks,colorlinks,citecolor=eccvblue]{hyperref}

\usepackage{orcidlink}

\begin{document}

\title{FSD-BEV: Foreground Self-Distillation for Multi-view 3D Object Detection} 

\titlerunning{Foreground Self-Distillation for Multi-view 3D Object Detection}

\author{Zheng Jiang\inst{1, 3, 4\star}\orcidlink{0009-0002-9961-0951} \and
Jinqing Zhang\inst{2\star}\orcidlink{0000-0001-7423-9266} \and
Yanan Zhang\inst{2}\orcidlink{0000-0003-1592-1067} \and 
Qingjie Liu\inst{1, 2, 5\dag}\orcidlink{0000-0002-5181-6451} \and \\
Zhenghui Hu\inst{1, 2\dag}\orcidlink{0000-0002-6106-0416} \and
Baohui Wang\inst{3} \and
Yunhong Wang\inst{1, 2}\orcidlink{0000-0001-8001-2703}}

\authorrunning{Z.~Jiang, J. Zhang et al.}

\institute{Hangzhou Innovation Institute, Beihang University, Hangzhou, China \and 
State Key Laboratory of Virtual Reality Technology and Systems, \\ Beihang University, Beijing, China \and
School of Software, Beihang University, Beijing, China \and
Shanghai ZEEKR Blue New Energy Technology Co., Ltd. \and Zhongguancun Laboratory, Beijing, China \\
{\tt \{jzzzz, zhangjinqing, zhangyanan, qingjie.liu\}@buaa.edu.cn, \\zhenghuihu2013@163.com, \{wangbh, yhwang\}@buaa.edu.cn} \\
}
\renewcommand{\thefootnote}{}
\footnotetext[2]{$\star$ Equal contribution.\quad$\dag$ Corresponding author.}
\maketitle

\begin{abstract}
Although multi-view 3D object detection based on the Bird's-Eye-View (BEV) paradigm has garnered widespread attention as an economical and deployment-friendly perception solution for autonomous driving, there is still a performance gap compared to LiDAR-based methods. In recent years, several cross-modal distillation methods have been proposed to transfer beneficial information from teacher models to student models, with the aim of enhancing performance. However, these methods face challenges due to discrepancies in feature distribution originating from different data modalities and network structures, making knowledge transfer exceptionally challenging. In this paper, we propose a Foreground Self-Distillation (FSD) scheme that effectively avoids the issue of distribution discrepancies, maintaining remarkable distillation effects without the need for pre-trained teacher models or cumbersome distillation strategies. Additionally, we design two Point Cloud Intensification (PCI) strategies to compensate for the sparsity of point clouds by frame combination and pseudo point assignment. Finally, we develop a Multi-Scale Foreground Enhancement (MSFE) module to extract and fuse multi-scale foreground features by predicted elliptical Gaussian heatmap, further improving the model's performance. We integrate all the above innovations into a unified framework named FSD-BEV. Extensive experiments on the nuScenes dataset exhibit that FSD-BEV achieves state-of-the-art performance, highlighting its effectiveness. \textit{The code and models are available at: \url{https://github.com/CocoBoom/fsd-bev}.}
\end{abstract}

\section{Introduction}
\label{sec:intro}

3D object detection is a fundamental task in autonomous driving. Although LiDAR-based methods \cite{zhang2021pc,zhou2018voxelnet,yin2021center,zhou2023octr} have demonstrated excellent performance, their limitations, such as the absence of color and texture information and the high hardware costs involved, have sparked a growing interest in approaches utilizing multi-view cameras. Early multi-view camera-based methods \cite{wang2022probabilistic,wang2021fcos3d,park2021pseudo} initially performed object detection tasks independently on individual images, followed by cross-view integration. However, this approach often resulted in suboptimal performance and hindered the coherence of subsequent tasks. Subsequent research has shifted focus towards the Bird’s-Eye-View (BEV) paradigm, offering a stronger universal intermediate representation for various views and modalities, and facilitating the integration of temporal information.

\begin{figure}[t]
  \centering
  \begin{subfigure}{0.4\linewidth}
    \centering
    \includegraphics[height=3.3cm]{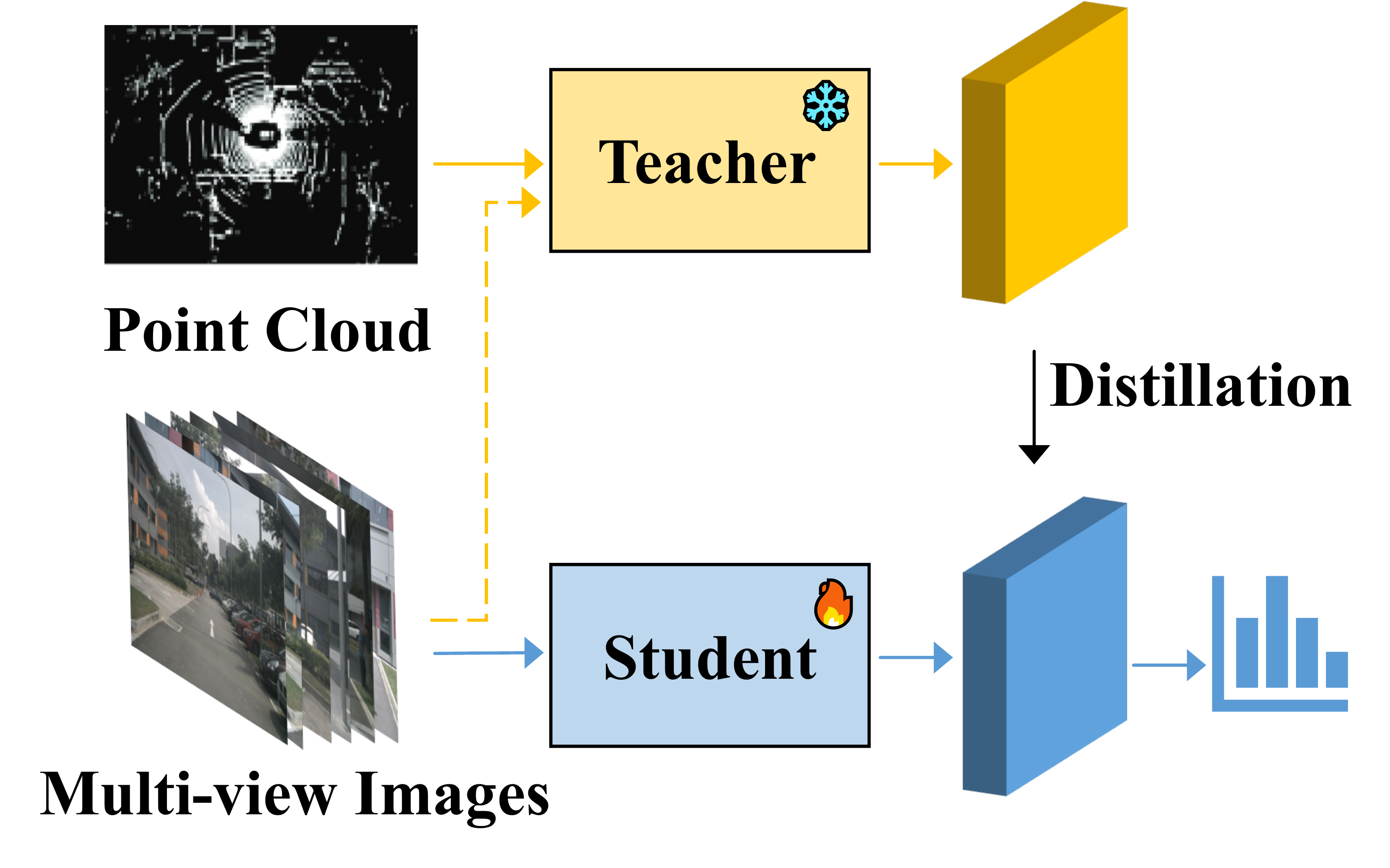}
    \caption{Cross-modal distillation}
    \label{Fig.2a}
   \end{subfigure}
   \begin{subfigure}{0.59\linewidth}
    \centering
    \includegraphics[height=3.3cm]{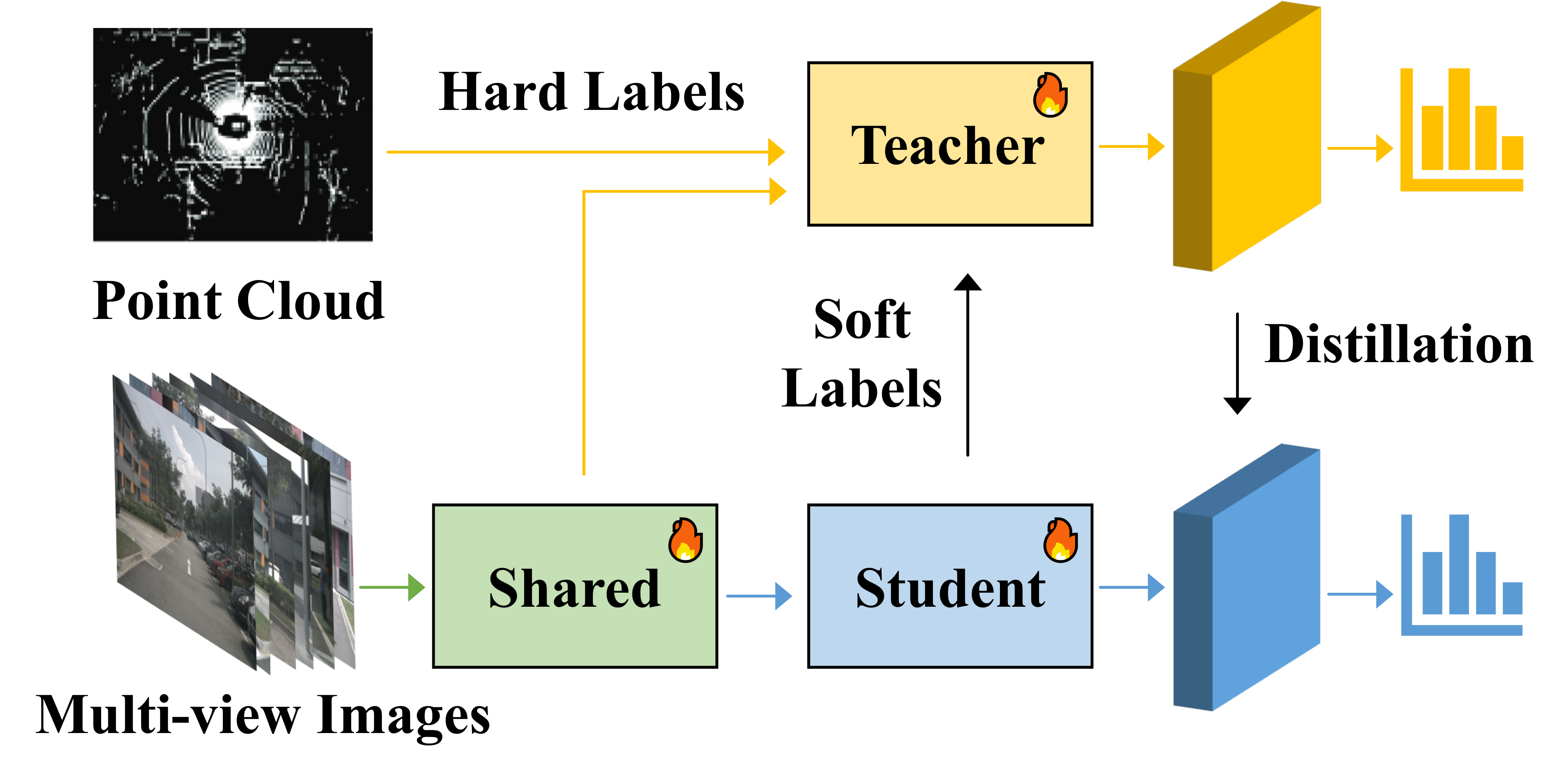}
    \caption{Self-distillation (ours)}
    \label{Fig.2b}
   \end{subfigure}
  \caption{Comparison of cross-modal distillation framework and our self-distillation framework. The hard labels denote the depth maps and foreground segmentation generated by LiDAR point clouds, the soft labels are those predicted by the student. The teacher branch and student branch share the same image features and thus mitigate the distribution discrepancies between the distillation targets.}
  \label{Fig.2}
\end{figure}

Although existing BEV-based multi-view 3D object detection methods have achieved competitive precision, there are still certain performance gaps compared to methods based on LiDAR. Distillation methods are employed to mitigate the performance gap between different modalities. Most of the BEV-based distillation methods~\cite{chen2022bevdistill,wang2023distillbev,zhou2023unidistill,huang2022tig} fall into the framework illustrated in \cref{Fig.2a}. A pre-trained teacher model transforms the LiDAR point clouds or multi-modal inputs into the frozen teacher BEV features, which serve as the inflexible guidance for the BEV features generated by the student. However, the distribution discrepancies between the distillation targets are the shared challenge encountered by these distillation methods, which are caused by differences in data modality, network structures and other factors. It makes it necessary to adopt cumbersome distillation strategies~\cite{wang2023distillbev,zhou2023unidistill,huang2022tig} to ensure the effectiveness.

In this paper, we propose the foreground self-distillation scheme for the BEV-based 3D objects, aiming to enhance model performance through the collaborative growth of teacher and student branches. Unlike the previous cross-modal distillation methods, a separate teacher model is not required in our framework; instead, a teacher branch is attached to the student model. As illustrated in \cref{Fig.2b}, the teacher branch utilizes the hard labels generated by LiDAR point clouds to obtain high-quality teacher BEV features and offer guidance to the student branch, while conversely, the student compensates the teacher with the predicted soft labels to fill the vacancy of the hard labels. It is worth noting that the teacher and student BEV features share the same source of image features in our self-distillation framework. It avoids severe distribution discrepancies and allows joint progress of the student and the teacher, which raises the upper bound on the distillation effect. Another challenge when distilling BEV features is that a large ratio of background provides trivial benefits and even disturbs the alignment in the foreground. Other methods commonly design complicated strategies to pay more attention to the foreground during the distillation process. In our scheme, the same purpose can be easily achieved by filtering out the background in both teacher and student BEV features following the way of SA-BEV~\cite{zhang2023sa}. We name the whole framework described above Foreground Self-Distillation (FSD).

The performance of the teacher branch in FSD heavily depends on the quality of the hard labels generated by point clouds. As a result, we design two Point Cloud Intensification (PCI) strategies to address the sparsity of point cloud data, namely, merging frame information and assigning pseudo points to objects that have no associated points. The sparsity of the hard labels is well mitigated, providing better guidance for feature synthesis. Furthermore, we develop a Multi-Scale Foreground Enhancement (MSFE) module to extract and fuse multi-scale foreground features by predicted elliptical Gaussian heatmap. This module effectively helps the student branch to predict more precise soft labels and thus improve the performance of the whole framework.

The strengths of our foreground self-distillation approach are: (1) No need for additional auxiliary models and pre-training processes; (2) Without cumbersome distillation strategies; (3) Avoidance of severe distribution discrepancies between the distillation targets; (4) Joint progress of the student and teacher model. We integrate the above designs into one framework named FSD-BEV. Extensive experiments validate the effectiveness of our method, demonstrating outstanding performance. In summary, the major contributions of this paper are:
\begin{itemize}
  \item We propose the first Foreground Self-Distillation framework for BEV-based multi-view 3D object detection, effectively mitigating the distribution discrepancies of the distilled features and background noise interference inherent in cross-modal distillation.
  \item We design two Point Cloud Intensification strategies to compensate for the sparsity of point clouds by frame combination and pseudo point assignment, further intensifying the performance ceiling of the teacher model.
  \item We develop a Multi-Scale Foreground Enhancement module to extract and fuse multi-scale foreground features by predicted elliptical Gaussian heatmap, further improving the model's performance.
\end{itemize}

\section{Related Work}

\subsection{Multi-view 3D Object Detection}
Multi-view 3D object detection is a critical task for autonomous driving systems, aiming to harness information from various perspectives. The Bird's-Eye View (BEV) representation, offering a unified representation of multiple views, divides research in this field into two paradigms: attention-based and depth-based. DETR3D \cite{wang2022detr3d} is a representative of the attention-based paradigm, utilizing sparse 3D queries to index 2D features extracted from images. BEVFormer \cite{li2022bevformer} introduced temporal sequences and incorporated spatiotemporal attention mechanisms. BEVFormerV2 \cite{yang2023bevformer} extended its focus to the gains from 2D tasks. The PETR series \cite{liu2022petr,liu2023petrv2,wang2023exploring} introduced 3D positional information to queries and underwent various improvements. 

In another branch, BEVDet \cite{huang2021bevdet} is the pioneer that combines LSS \cite{philion2020lift} with the detection head, treating depth information as an explicit prediction for detection tasks. BEVDepth \cite{li2023bevdepth} uses LiDAR ground truth as depth supervision, significantly enhancing detection performance. BEVStereo \cite{li2023bevstereo} further refines depth predictions by utilizing stereo vision matching. STS \cite{wang2022sts} enhances depth quality by leveraging geometric correspondences between adjacent frames. However, regardless of the type of method, existing multi-view 3D object detection still lags behind LiDAR-based methods due to insufficiently acquired geometric information. In this paper, we focus on designing more suitable distillation techniques to narrow the performance gap between multi-view 3D object detection and LiDAR-based methods.

\subsection{Knowledge Distillation}
Knowledge distillation is employed for capacity learning between two models. It enhances the performance of the model without altering the student model structure and has been widely applied \cite{dai2021general,guo2021distilling,hou2020inter}. Self-distillation \cite{zhang2019your,yang2019snapshot} is another approach that implements distillation between features at different levels or transferring knowledge across various periods. In the field of 3D object detection, LIGA-stereo \cite{guo2021liga} utilizes LiDAR-based detectors to guide Stereo-based detectors. UVTR \cite{li2022unifying} unifies multi-modal representations in voxel space, enabling accurate single-modal or cross-modal 3D detection. BEVDistill \cite{chen2022bevdistill} performs cross-modal distillation in a unified BEV space. UniDistill \cite{zhou2023unidistill} proposes three distillation approaches in three stages. DistillBEV \cite{wang2023distillbev} proposes a balanced design for the fusion of multi-scale and temporal information. However, existing methods generally require additional training of teacher models and suffer from the divergence in modality and model structure between teacher and student, resulting in limited performance gains from knowledge distillation. In this paper, we propose a foreground self-distillation scheme that effectively alleviates distribution discrepancies, achieving excellent distillation effects without additional pre-trained models.

\begin{figure*}[t] 
  \centering 
    \includegraphics[width=1\linewidth]{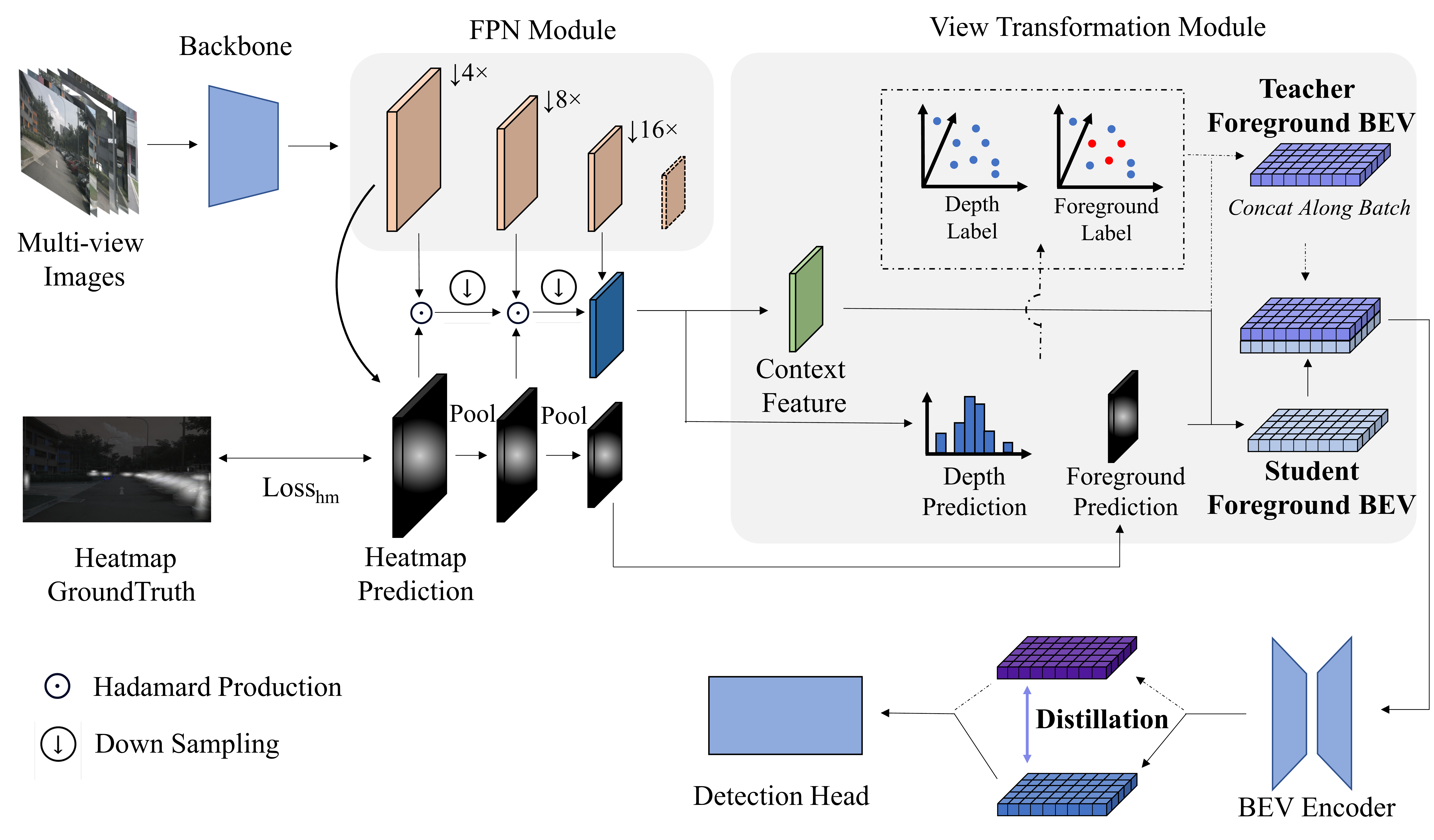}
    \caption{The overall architecture of the proposed FSD-BEV. The features enhanced by the foreground heatmap are fed into the View Transformation Module to generate the student BEV. The teacher branch generates the teacher BEV by combining hard labels with soft labels from the student branch. Subsequently, they are concatenated along the batch dimension for subsequent joint training and undergo distillation operations before entering the detection head.} 
  \label{Fig.framework} 
\end{figure*}

\section{Method}

In this section, we will present the details of FSD-BEV, the first self-distillation approach for the BEV-based 3D detectors. The overall architecture of FSD-BEV is illustrated in~\cref{Fig.framework}. Unlike the former BEV-based cross-modal distillation methods that employ extra pre-trained teacher models and use the cumbersome distillation strategy, Foreground Self-Distillation achieves significant accuracy improvement by simply doing feature alignment within a single model. Subsequently, we conduct several Point Cloud Intensification procedures to build more truthful teacher BEV features and further improve the effect of distillation. Finally, we effectively aggregate foreground information from different scale feature maps using the Multi-Scale Foreground Enhancement strategy, extracting more useful 2D image features.

\begin{figure}[t]
  \centering
  \includegraphics[width=0.8\linewidth]{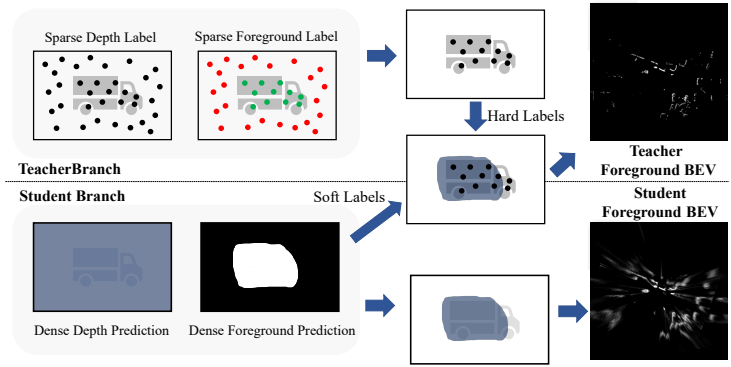}
  \caption{Details of BEV feature generation. Due to the sparse nature of the depth label, the teacher branch cannot acquire dense foreground information. To address this limitation, we use soft labels generated by the student branch to assist the teacher branch in generating denser depth maps.}
  \label{Fig.3.1}
\end{figure}

\subsection{Foreground Self-Distillation}
\label{subsec.3.1} 

The former BEV-based cross-modal distillation methods \cite{chen2022bevdistill,zhou2023unidistill,huang2022tig,wang2023distillbev} generally require a heavier LiDAR-based or multi-modal teacher model to serve as the guidance. And BEV features are commonly employed by these methods as the distillation medium. However, the distribution gap between the BEV features of teacher and student features poses challenges in cross-modal distillation. Such a distribution gap may arise from differences in data modality, network structures, and other factors. The focus of distillation methods lies in transferring features effectively under distribution disparities, necessitating the design of clever strategies to maximize distillation efficiency. Since the imitation in the background region offers trivial benefits to the detection accuracy, it is a natural idea to select foreground regions by projecting ground truths boxes onto the BEV and perform delicate distillation operations there.

In the proposed Foreground Self-Distillation, significant performance improvement is achieved without the need for additional auxiliary models or elaborate distillation strategies. Unlike employing pre-trained teacher models in conventional distillation approaches, the teacher and student engage in a joint learning process in our method. While the teacher and student BEV features share the same context feature resources, more accurate depth and semantic information help construct the high-performing teacher BEV features, providing ongoing guidance for the student BEV features. The foreground segmentation is also introduced to generate BEV features that only contain foreground information, which abandons the useless imitation of the background region and avoids noise interference. The foreground segmentation also dramatically increases the accuracy of the teacher branch, consequently strengthening the distillation efficiency.

\subsubsection{Student BEV} 

As illustrated in \cref{Fig.3.1}, we generate the student BEV features in the View Transformation Module. Following conventional procedures \cite{li2023bevdepth}, we predict context features $\bold{C}$ and predicted depth maps $\bold{D}$ from the input image feature map. To filter out the background in the BEV features, the predicted foreground segmentation $\bold{S}$ is required. It can be obtained by reusing the DepthNet designed for predicting $\bold{D}$ or by the specialized module described in~\cref{subsec.3.3}. We leverage SA-BEVPool \cite{zhang2023sa} to generate BEV features that contain only foreground information. The student BEV features can be represented by:

\begin{equation}
\bold{B}_{s} = \text{SA-BEVPool}( \bold{C}, \bold{D}, \bold{S}) ,
\end{equation}
which means $\bold{C}$, $\bold{D}$ and $\bold{S}$ are taken as the input of SA-BEVPool and the student BEV features carrying only foreground information are created.

\subsubsection{Teacher BEV} 

In another branch of \cref{Fig.3.1}, we generate the teacher BEV features by injecting accurate depth and semantic information. Specifically, we utilize the ground truth depth maps $\hat{\bold{D}}$ and foreground segmentation $\hat{\bold{S}}$, obtained from LiDAR point clouds as stated in SA-BEV~\cite{zhang2023sa}, to replace the relatively unfaithful $\bold{D}$ and $\bold{S}$. While naming the ground truth as the hard labels, the predicted $\bold{D}$ and $\bold{S}$ are known as the soft labels. Although hard labels reveal accurate scene information, the sparsity of point clouds can lead to the label missing. To address this issue, soft labels are utilized to fill the missing part of the hard labels and the combined labels can be represented by:
\begin{align}
    \bar{\bold{D}} &= \bold{M}\hat{\bold{D}} + (1-\bold{M})\bold{D} \\
    \bar{\bold{S}} &= \bold{M}\hat{\bold{S}} + (1-\bold{M})\bold{S},
\end{align}
where $\bold{M}$ denotes the valid mask of the hard labels. The value of $\bold{M}$ is 1 where the hard labels are available and 0 otherwise. The teacher BEV features are then generated by:
\begin{equation}
\bold{B}_{t} = \text{SA-BEVPool}( \bold{C}, \bar{\bold{D}}, \bar{\bold{S}}).
\end{equation}

Since both student and teacher BEV features inherit the information of $\bold{C}$ in our self-distillation framework, the gap between them is much smaller than in the former cross-modal distillation framework. Meanwhile, part of the soft labels predicted by the student branch are also involved in constructing teacher BEV features, making it easier for the student to imitate the teacher. 

\subsubsection{Collaborative Training} 

It is not easy to directly align $\bold{B}_{s}$ and $\bold{B}_{t}$. The former distillation methods attach additional adaptation modules, which map the student BEV features to the teacher BEV features. Instead of following the conventional approach, we concatenate the $\bold{B}_{s}$ and $\bold{B}_{t}$ together along the batch dimension. Subsequently, they jointly undergo processing through the BEV encoder, resulting in high-level BEV features, which can be represented by:
\begin{align}
    \hat{\bold{B}}_{s} &= \text{BEVEncoder}(\bold{B}_{s}) \\
    \hat{\bold{B}}_{t} &= \text{BEVEncoder}(\bold{B}_{t}). 
\end{align}
The alignment is then applied between $\hat{\bold{B}}_{s}$ and $\hat{\bold{B}}_{t}$. It can be found that the BEV encoder plays the same role as the adaptation modules without increasing the parameters. The BEV encoder can also be regarded as a feature filter, which filters the incorrect information in $\bold{B}_{s}$ and makes the filtered $\hat{\bold{B}}_{s}$ similar to $\hat{\bold{B}}_{t}$.

It is noteworthy that $\hat{\bold{B}}_{s}$ and $\hat{\bold{B}}_{t}$ should also be concatenated and jointly fed into the detection head. If $\hat{\bold{B}}_{t}$ are not supervised by the detection losses, it will move to $\hat{\bold{B}}_{s}$ after the alignment, which is opposite to our purpose. When $\hat{\bold{B}}_{s}$ and $\hat{\bold{B}}_{t}$ are simultaneously supervised, the relative unfaithful $\hat{\bold{B}}_{s}$ tends to incur larger detection losses than the $\hat{\bold{B}}_{t}$, which mean the participation of $\hat{\bold{B}}_{t}$ will not heavily influence the training of student branch. 

\subsubsection{Distillation Loss}

Benefiting from the homogeneity between $\hat{\bold{B}}_{s}$ and $\hat{\bold{B}}_{t}$ and their foreground-only characteristic, we circumvent the necessity of devising intricate distillation losses for ensuring effective information transfer. Instead, the simple ${L_2}$ loss can suffice. A problem is that both $\hat{\bold{B}}_{s}$ and $\hat{\bold{B}}_{t}$ are not frozen during collaborative training, which means the ${L_2}$ distillation loss can take a shortcut by decreasing the magnitudes of $\hat{\bold{B}}_{s}$ and $\hat{\bold{B}}_{t}$ instead of pulling them together. As a result, we normalize the magnitudes of $\hat{\bold{B}}_{s}$ and $\hat{\bold{B}}_{t}$ before calculating the distillation loss. The distillation loss can be represented by: 
\begin{equation}
  \mathcal{L}_{distill} = \frac{1}{{H \times W}}\sum\limits_i^H {\sum\limits_j^W {{{\left\| \frac{\hat{\bold{B}}_{t}^{ij}-\hat{\bold{B}}_{s}^{ij}}{\|\hat{\bold{B}}_{t}^{ij}\|_2}\right\|_2}}} }.
\end{equation}

\begin{figure}[t]
  \centering
  \begin{subfigure}{0.3\linewidth}
    \centering
    \includegraphics[width=0.8\linewidth]{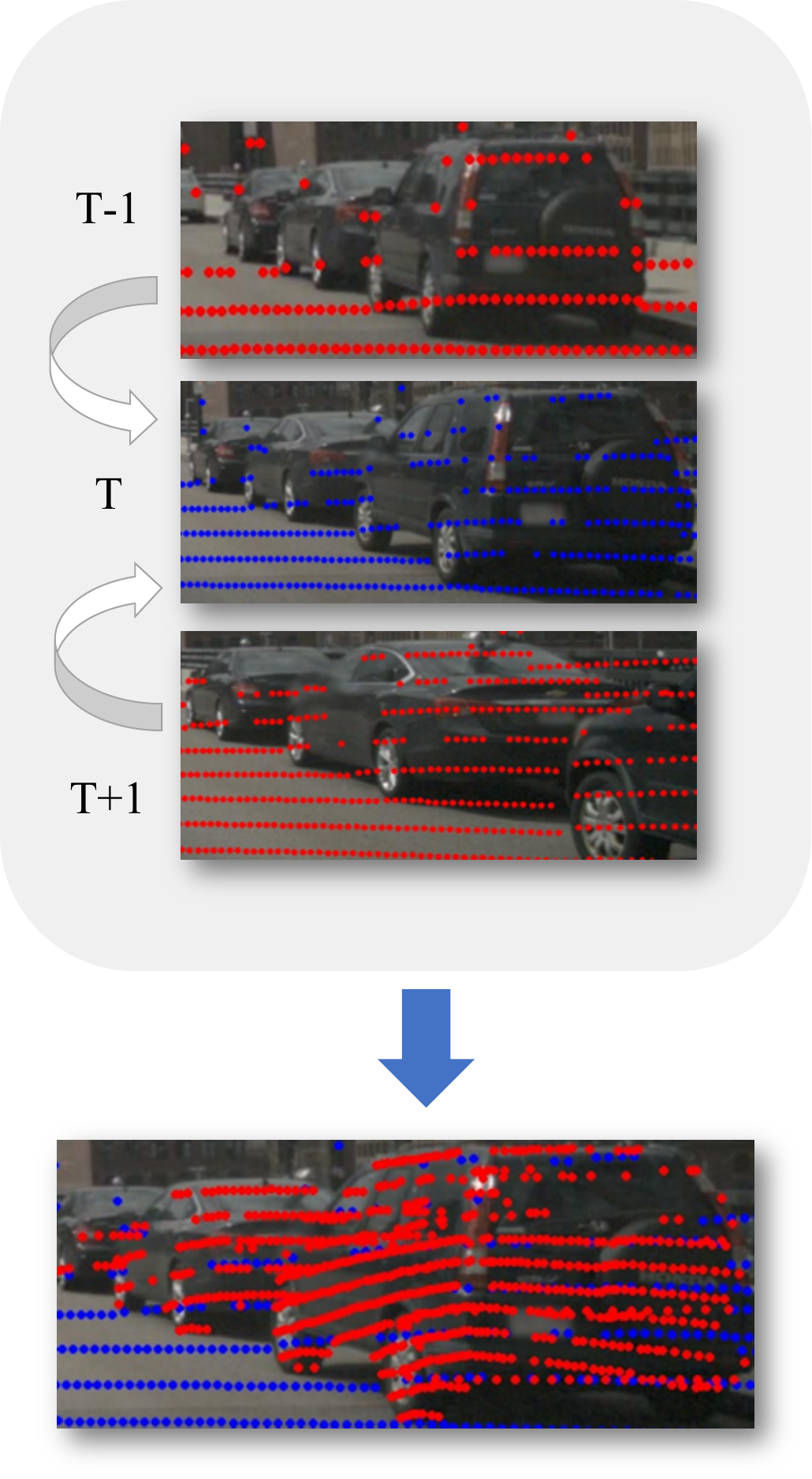}
    \caption{Frame Combination}
    \label{Fig.3.2a}
   \end{subfigure}
   \begin{subfigure}{0.6\linewidth}
    \centering
    \includegraphics[width=0.8\linewidth]{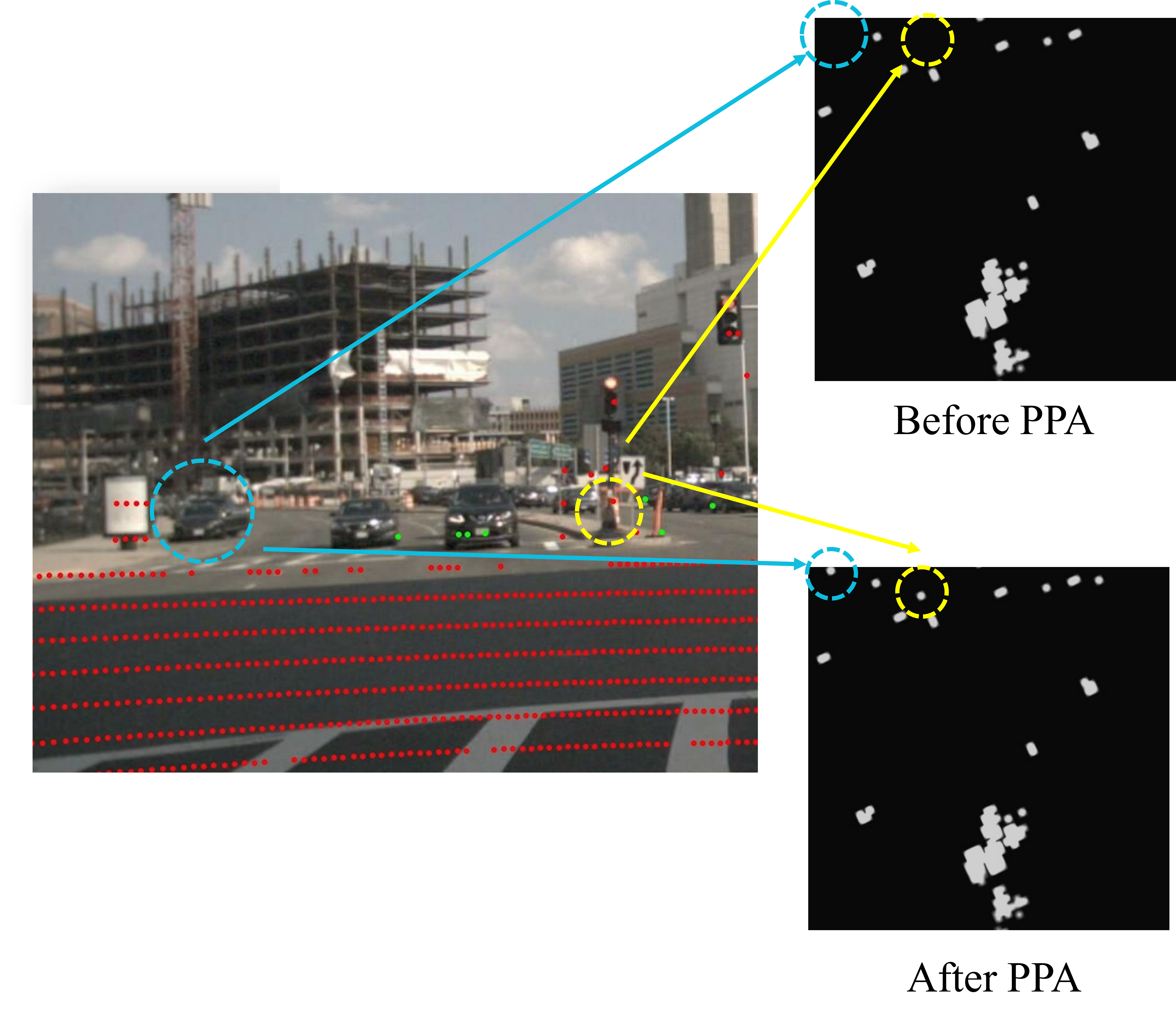}
    \caption{Pseudo Point Assignment}
    \label{Fig.3.2b}
   \end{subfigure}
  \caption{Overview of Point Cloud Intensification. In (a), we enhance the current frame by merging points of static objects from the adjacent frames in time. In (b), green and red points represent foreground and background points, respectively. We employ the PPA strategy to complete missing point clouds for foreground objects.}
  \label{Fig.3.2}
\end{figure}

\subsection{Point Cloud Intensification}
\label{subsec.3.2}

The quality of hard labels $\hat{\bold{D}}$ and $\hat{\bold{S}}$ generated by LiDAR point clouds determines the performance of the teacher branch, thereby influencing the distillation results. However, the sparsity of point clouds often results in plenty of distant objects having few or no points, weakening the quality of hard labels. We introduce two kinds of Point Cloud Intensification (PCI) strategies here to make hard labels carry more information about the scene as shown in~\cref{Fig.3.2}.

\subsubsection{Frame Combination}

The first PCI strategy named Frame Combination (FC) uses the adjacent frames in time to provide supplementary point clouds. To avoid the points of the dynamic objects introducing errors, only the points belonging to the stationary foreground objects will be combined, such as the parked cars, the cycles without riders and the traffic cones. We transform the point clouds of the adjacent frames into the coordinate system of the current frame and use the stationary object boxes to select the desired points. the objects in the current frame will have denser points after FC as illustrated in \cref{Fig.3.2a}.

\subsubsection{Pseudo Point Assignment}

After combining the adjacent point clouds, there are still some objects that do not show up on the hard labels. They could be dynamic objects that are not suitable for Frame Combination, or be quite distant that even the adjacent frames can not provide valid points. In this situation, it is a reasonable choice to assign the approximate points to these objects in the space, which is another PCI strategy called Pseudo Point Assignment (PPA).

We first project the ground truth 3D boxes onto the image to get their corresponding 2D boxes. Each 2D box can be represented by $( x_{1}  , y_{1} , x_{2}, y_{2})$, where $(x_{1}, y_{1})$ denotes the top-left point, $(x_{2}, y_{2})$ denotes the bottom-right point. We select the boxes that should be assigned with a pseudo point following several criteria: 1) there are no real points within the box after Frame Combination; 2) the depth of the box is in the perception range; 3) the box has good visibility (set to 3 or 4 in the nuScenes \cite{caesar2020nuscenes} dataset). The coordinate of the pseudo point in the image coordinate system can be represented by: 
\begin{equation}
  p_{pseudo} = ( \frac{x_{1} + x_{2}}{2}, \frac{y_{1} + y_{2}}{2}, d_{corner} ),
\end{equation}
where $d_{corner}$ represents the minimum depth among the eight corner points of the 3D box, which is close to the surface depth of the object. As shown in~\cref{Fig.3.2b}, the teacher BEV features can capture the missing objects after PPA.

\subsection{Multi-Scale Foreground Enhancement}
\label{subsec.3.3} 

Fusing high-scale features is a straightforward way to provide finer depth maps to the view transformation module. Meanwhile, we think the foreground is more beneficial than the background in high-scale features. We introduce a Multi-Scale Foreground Enhancement (MSFE) module as shown in~\cref{Fig.framework} to facilitate the integration of extensive foreground details into the model's learning paradigm.

We define 
${\bold{F}^{4}, \, \bold{F}^{8}, \, \bold{F}^{16} }$ 
as the output feature maps of the FPN \cite{lin2017feature} layer, where the superscript represents the downsampling ratio of the feature map size from the input image size. Unlike the former BEV-based methods commonly utilize $\bold{F}^{16}$ as the input features for the depth network, we predict the high-scale foreground segmentation $\bold{S}^4$ by using $\bold{F}^{4}$ to provide more detailed information.  

On such a high scale, the labels generated by LiDAR point clouds are fairly sparse (about 80\% of the labels are vacant) and cannot offer effective supervision. Instead, we adopt the heatmaps generated by the ground truth 2D boxes on the images as the labels. Unlike CenterNet \cite{zhou2019objects} drawing circular Gaussian distributions around the object centers, we represent the confidence distribution of foreground objects by elliptical Gaussians that fill the 2D boxes, and longer dimension has larger covariance. Focal Loss is utilized as the training loss.

Once $\bold{S}^4$ is predicted, we gather the foreground information of multi-scale features and integrate them through several dot product and downsampling operations. 
The $\bold{S}^4$ is first filtered by a threshold $\beta$ to get $\bold{S}^4_f$, which can be represented by:
\begin{equation}
\bold{S}^{4}_f=M_{f} \cdot \bold{S}^{4}, \; M_{f}= \begin{matrix}
\end{matrix}\begin{cases}
  1,& \bold{S}^{4}\ge \beta    \\
  0,& \bold{S}^{4}< \beta
\end{cases}.
\end{equation}
Denoting $\text{DS}^n$ as the downsampling operation by a factor of $n$, the enhanced $\bold{F}^{16}_{MSFE}$ can be represented by:
\begin{equation}
\bold{F}^{16}_{MSFE} = \bold{F}^{16} + \text{DS}^2(\bold{F}^{8}\cdot\text{DS}^2(\bold{S}^{4}_f)) + \text{DS}^4(\bold{F}^{4}\cdot\bold{S}^{4}_f)
\end{equation}
After feature aggregation, $\bold{F}^{16}_{MSFE}$ is supplemented with more detailed information, making the prediction based on it more precise. The high-scale $\bold{S}^4$ can also be utilized by the student branch to offer a finer foreground segmentation.


\section{Experiments}

\subsection{Experimental Settings}
\subsubsection{Dataset and Metrics} 
We conducted our experiments on the nuScenes \cite{caesar2020nuscenes} dataset, which comprises 1000 sequences captured by sensors such as cameras and LiDAR. Each sequence is approximately 20 seconds long, with 700 sequences used for training, 150 for validation, and 150 for testing. We employed official evaluation metrics, including mAP, ATE, ASE, AOE, AVE, and AAE, to assess the results for center distance, translation, scale, orientation, velocity, and attributes, respectively. Additionally, the official evaluation provides a comprehensive metric known as nuScenes Detection Score (NDS). 

\subsubsection{Implementation Detail}

Our model design follows the classical depth-based BEV detection model architecture. We conducted experiments using 8 NVIDIA GeForce RTX 3090 GPUs. AdamW \cite{loshchilov2017decoupled} is employed as our optimizer with a learning rate set to 2e-4 and kept constant throughout the training process. Additionally, we adopted mixed precision training to save more resources. In most ablation experiments, we utilized ResNet50 as the backbone network with an image input size of 256×704. When comparing with other state-of-the-art models, we use larger image size and adopt heavier backbones. If not stated, the model is trained for 24 epochs with the CBGS \cite{zhu2019class} strategy. Our experiments are conducted using only one past frame with an interval of 0.5s. 

\begin{table}[t]

\centering
\caption{Comparison with the state-of-the-art methods on the nuScenes ${val}$ set. ${^{*}}$ Benefited from the perspective-view pre-training. ${^{\dag}}$ Increasing the size of BEV features from 128×128 to 256×256 to enhance the granularity of distillation.} 
\label{tab:val}
\renewcommand\arraystretch{1.2}

\setlength{\tabcolsep}{1mm}{
\resizebox{\linewidth}{!}{
\begin{tabular}{l|cc|c|cc|ccccc}
\hline
Methods & Backbone & Resolution & Frames & mAP↑ & NDS↑ & mATE↓ & mASE↓ & mAOE↓ & mAVE↓ & mAAE↓ \\ 
\hline
BEVDet~\cite{huang2021bevdet} & ResNet50 & 256$\times$704 & 1 & 0.298 & 0.379 & 0.725 & 0.279 & 0.589 & 0.860 & 0.245\\
BEVDet4D\cite{huang2022bevdet4d}       & ResNet50 & 256$\times$704 & 2 & 0.322 & 0.457 & 0.703 & 0.278 & 0.495 & 0.354 & 0.206 \\
PETRv2\cite{liu2023petrv2}       & ResNet50 & 256$\times$704 & 2 & 0.349 & 0.456 & 0.700 & 0.275 & 0.580 & 0.437 & \textbf{0.187} \\
BEVDepth\cite{li2023bevdepth}       & ResNet50 & 256$\times$704 & 2 & 0.351 & 0.475 & 0.639 & 0.267 & 0.479 & 0.428 & 0.198 \\
BEVStereo \cite{li2023bevstereo} & ResNet50 & 256$\times$704 & 2 & 0.372 & 0.500 & 0.598 & 0.270 & 0.438 & 0.367 & 0.190\\
FB-BEV \cite{li2023fb} & ResNet50 & 256$\times$704 & 3 & 0.378 & 0.498 & 0.620 & 0.273 & 0.444 &0.374 & 0.200 \\
SA-BEV\cite{zhang2023sa}       & ResNet50 & 256$\times$704 & 2 & 0.387 & 0.512 & 0.613 & 0.266 & \textbf{0.352} & 0.382 & 0.199 \\
\rowcolor{gray!20}
FSD-BEV       & ResNet50 & 256$\times$704 & 2 & 0.403 & 0.526 & 0.576 & 0.259 & 0.362 & 0.359 & 0.198 \\ 
\rowcolor{gray!20}
FSD-BEV$^{\dag}$   & ResNet50 & 256$\times$704 & 2 & \textbf{0.412} & \textbf{0.538} & \textbf{0.527} & \textbf{0.256} & 0.363 & \textbf{0.330} & 0.207 \\ 
\hline
DETR3D\cite{wang2022detr3d}       & ResNet101 & 900$\times$1600 & 1 & 0.303 & 0.374 & 0.860 & 0.278 & 0.437 & 0.967 & 0.235 \\
PETR\cite{liu2022petr}${^{*}}$         & ResNet101 & 512$\times$1408 & 1 & 0.366 & 0.441 & 0.717 & 0.267 & 0.412 & 0.834 & 0.190 \\
BEVFormer\cite{li2022bevformer}${^{*}}$    & ResNet101 & 900$\times$1600 & 4 & 0.416 & 0.517 & 0.673 & 0.274 & 0.372 & 0.394 & 0.198 \\
BEVDepth\cite{li2023bevdepth}     & ResNet101 & 512$\times$1408 & 2 & 0.412 & 0.535 & 0.565 & 0.266 & 0.358 & 0.331 & 0.190 \\
Sparse4D\cite{lin2022sparse4d}${^{*}}$  & ResNet101 & 900$\times$1600 & 4 & 0.436 & 0.541 & 0.633 & 0.279 & 0.363 & 0.317 & \textbf{0.177} \\
TiG-BEV\cite{huang2022tig} & ResNet101 & 512$\times$1408 & 2 & 0.440 & 0.544 & 0.570 & 0.267 & 0.392 & 0.331 & 0.201 \\
STS~\cite{wang2022sts} & ResNet101 & 512$\times$1408 & 2 & 0.431 & 0.542 & 0.525 & 0.262 & 0.380 & 0.369 & 0.204 \\
StreamPETR\cite{wang2023exploring}${^{*}}$       & ResNet101 & 512$\times$1408 & 8 & \textbf{0.504} & 0.592 & 0.569 & 0.262 & \textbf{0.315} & \textbf{0.257} & 0.199 \\
\rowcolor{gray!20}
FSD-BEV       & ResNet-101 & 512$\times$1408 & 2 & 0.488 & 0.589 & 0.501 & \textbf{0.251} & 0.316 & 0.288 & 0.197 \\
\rowcolor{gray!20}
FSD-BEV${^{*}}$       & ResNet-101 & 512$\times$1408 & 2 & 0.500 & \textbf{0.596} & \textbf{0.489} & \textbf{0.251} & 0.317 & 0.279 & 0.204 \\ 
\hline
\end{tabular}}}
\end{table}

\begin{table*}[t]

\centering
\caption{Comparison with the state-of-the-art methods on the nuScenes ${test}$ set.} 
\label{tab:test}
\renewcommand\arraystretch{1.2}

\setlength{\tabcolsep}{1mm}{
\resizebox{\linewidth}{!}{
\begin{tabular}{l|cc|c|cc|ccccc}
\hline
Methods & Backbone & Resolution & Frames & mAP↑ & NDS↑ & mATE↓ & mASE↓ & mAOE↓ & mAVE↓ & mAAE↓  \\ 
\hline
DETR3D\cite{wang2022detr3d}       & V2-99 & 900$\times$1600 & 1 & 0.412 & 0.479 & 0.641 & 0.255 & 0.394 & 0.845 & 0.133 \\
UVTR\cite{li2022unifying}       & V2-99 & 900$\times$1600 & 1 & 0.472 & 0.551 & 0.577 & 0.253 & 0.391 & 0.508 & \textbf{0.123} \\
BEVFormer\cite{li2022bevformer}    & V2-99 & 900$\times$1600 & 4 & 0.481 & 0.569 & 0.582 & 0.256 & 0.375 & 0.378 & 0.126 \\
PETRv2\cite{liu2023petrv2}  & RevCol-L & 640$\times$1600 & 2 & 0.512 & 0.592 & 0.547 & 0.242 & 0.360 & 0.367 & 0.126 \\
BEVDepth\cite{li2023bevdepth}     & ConvNeXt-B & 640$\times$1600 & 2 & 0.520 & 0.609 & 0.445 & 0.243 & 0.352 & 0.347 & 0.127 \\
BEVStereo~\cite{li2023bevstereo} & V2-99 & 640$\times$1600 & 2 & 0.525 & 0.610 & 0.431 & 0.246 & 0.358 & 0.357 & 0.138\\
SA-BEV\cite{zhang2023sa}     & V2-99 & 640$\times$1600 & 2 & 0.533 & 0.624 & 0.430 & 0.241 & 0.338 & 0.282 & 0.139 \\
SOLOFusion\cite{park2022time}       & ConvNeXt-B & 640$\times$1600 & 17 & 0.540 & 0.619 & 0.453 & 0.257 & 0.376 & \textbf{0.276} & 0.148 \\
\rowcolor{gray!20}
FSD-BEV       & V2-99 & 640$\times$1600 & 2 & \textbf{0.543} & \textbf{0.633} & \textbf{0.404} & \textbf{0.236} & \textbf{0.325} & 0.286 & 0.131 \\ 
\hline
\end{tabular}}
}
\end{table*}

\subsection{Main Results}

We compare our method with state-of-the-art approaches on the val and test sets of nuScenes \cite{caesar2020nuscenes}. As shown in \cref{tab:val}, when we use the smaller ResNet50 as the backbone network, our method outperforms all the methods using only two frames. 
Further performance gains are observed when we increase the size of distilled BEV features, which exceeds SA-BEV \cite{zhang2023sa} by 2.5\% and 2.6\% in mAP and NDS. When we use the larger ResNet101 as the backbone, the performance of FSD-BEV is further increased and even exceeds StreamPETR \cite{wang2023exploring}, the current state-of-the-art by 0.4\% NDS. It is noteworthy that StreamPETR uses much more past frames than FSD-BEV.

The comparison results on the test set are illustrated in \cref{tab:test}. Compared with SOLOFusion \cite{park2022time} that utilizes a larger ConvNeXt-B backbone network and inputs 17 past frames, FSD-BEV achieves a lead of 0.3\% and 1.4\% in mAP and NDS. When compared to the advanced method SA-BEV \cite{zhang2023sa}, which also uses 2 frames, the mAP and NDS of FSD-BEV are improved by 1\% and 0.9\%.

\begin{table}[t]
\centering
\caption{Performance comparison of different module combinations.}
\label{tab:ma}
\renewcommand\arraystretch{1.2}
\scriptsize
\setlength{\tabcolsep}{2mm}{
\begin{tabular}{ccc|cc|ccccc}
\hline
FSD & PCI & MSFE & mAP↑ & NDS↑ & mATE↓ & mASE↓ & mAOE↓ & mAVE↓ & mAAE↓\\ 
\hline
  &  &  & 0.363 & 0.486 & 0.620 & 0.276 & 0.450 & 0.387 & 0.222 \\
\checkmark  &  &  & 0.394 & 0.516 & 0.590 & 0.261 & 0.403 & 0.353 & 0.200 \\
\checkmark & \checkmark & & 0.400 & 0.516 & 0.586 & 0.261 & 0.436 & \textbf{0.349} & 0.202\\
\checkmark & \checkmark & \checkmark & \textbf{0.403} & \textbf{0.526} & \textbf{0.576} & \textbf{0.259} & \textbf{0.362} & 0.359 & \textbf{0.198}\\
\hline
\end{tabular}}
\end{table}
\subsection{Ablation Study}
\subsubsection{Module Analysis}
We select BEVDepth as the baseline and conduct training for 24 epochs using the CBGS strategy. We compare the performance of different module combinations. As shown in \cref{tab:ma}, when self-distillation is applied individually, significant improvements of 3.1\% mAP and 3.0\% NDS are achieved. After applying Point Cloud Intensification, there is an obvious improvement in mAP, indicating the enhanced capability of detecting difficult objects. It is consistent with the purpose of this module. When all three modules are integrated, the mAP and NDS are improved by 4\% and 4\% compared to the baseline. This fully demonstrates the effectiveness of each module.

\begin{table}[t]
\parbox{.47\linewidth}{
\centering
\caption{Ablation study of foreground segmentation applied in Foreground Self-Distillation. ``FS'' denotes the foreground segmentation.}
\label{tab:fsd}
\renewcommand\arraystretch{1}
\setlength{\tabcolsep}{2mm}{
\begin{tabular}{l|c|cc}
\hline
FS & Branch & mAP & NDS \\ 
\hline
  & Teacher & 0.468 & 0.564 \\
  & Student & 0.372 & 0.498 \\
\hline
\checkmark & Teacher & 0.584 & 0.641 \\
\checkmark & Student & 0.393 & 0.516 \\
\hline
\end{tabular}}}
\hfill
\parbox{.47\linewidth}{
\centering

\caption{Comparison of the Point Cloud Intensification schemes on the teacher branch. ``FC'' represents Frame Combination, and ``PPA'' represents Pseudo Point Assignment.}
\label{tab:pci}
\renewcommand\arraystretch{1}

\setlength{\tabcolsep}{2mm}{
\begin{tabular}{cc|cc}
\hline

FC & PPA& mAP & NDS \\ 
\hline
&  & 0.584 & 0.641 \\
\checkmark &  & \textbf{0.605} & 0.650 \\
\checkmark & \checkmark & \textbf{0.605} & \textbf{0.652} \\ 
\hline
\end{tabular}}}
\end{table}

\subsubsection{Foreground Self-Distillation}

We conduct experiments on Foreground Self-Distillation to highlight the benefits of distillation performance brought by foreground segmentation and the results are shown in \cref{tab:fsd}. It can be found that the student model applying the foreground segmentation outperforms the one without the foreground segmentation by 2.1\% mAP, 1.8\% NDS. It is partly because only distilling in the foreground region transfers more valuable information, and also because of the giant leap in the teacher's performance (11.6\% mAP and 7.7\% NDS), telling the importance of a good teacher.

\subsubsection{Point Cloud Intensification}
We separately evaluate the accuracy of the teacher branch to assess the impact of different Point Cloud Intensification strategies. As shown in \cref{tab:pci}, when employing the FC strategy, we observed improvements of 2.1\% in mAP and 0.9\% in NDS. However, upon further addition of the PPA strategy, only a slight increase in NDS is observed. This could be attributed to the fact that the FC method might possess similar functionality to PPA, resulting in a reduced gain from PPA.

\begin{table}[t]
\centering
\caption{Comparison of multi-scale foreground enhancement effects under different foreground thresholds. ${\beta}$ represents the foreground threshold, and ${\bold{F}^{4}}$ and ${\bold{F}^{8}}$ respectively denote feature maps of different scales (refer to \cref{subsec.3.3}).}
\label{tab:MSFE}
\renewcommand\arraystretch{1.2}
\scriptsize
\setlength{\tabcolsep}{2mm}{
\begin{tabular}{c|cc|cc|ccccc}
\hline
${\beta}$ & ${\bold{F}^{4}}$ & ${\bold{F}^{8}}$ & mAP↑ & NDS↑ & mATE↓ & mASE↓ & mAOE↓ & mAVE↓ & mAAE↓ \\ 
\hline
0.00&\checkmark&\checkmark&0.348&0.465&0.681&0.278&0.572&0.352&0.214\\
0.15&\checkmark&\checkmark&0.346&0.466&0.660&0.281&0.574&0.357&\textbf{0.202}\\ 
0.20&\checkmark&\checkmark&0.346&0.465&0.660&\textbf{0.276}&0.560&0.379&0.209 \\
0.30&\checkmark&\checkmark&0.344&0.460&0.687&0.278&0.594&0.359&0.207 \\
\hline
0.10&\checkmark&&0.347&0.467&0.677&0.281&\textbf{0.553}&0.350&0.207\\ 
0.10&&\checkmark&0.350&0.468&\textbf{0.659}&0.278&0.564&0.350&0.220\\ 
0.10&\checkmark&\checkmark&\textbf{0.352}&\textbf{0.469}&0.662&0.282&0.580&\textbf{0.340}&0.209\\ 
\hline
\end{tabular}}
\end{table}

\subsubsection{Multi-Scale Foreground Enhancement}
Different from pixel-level semantic segmentation tasks, the foreground masks suffice the requirement in FSD-BEV as long as they can roughly indicate the location of foreground objects. A high foreground threshold can filter out more background noise of the scene, but it also leads to possible losses of foreground information. Therefore, we compared the effects of foreground enhancement under different foreground threshold settings and different scale feature maps, as shown in \cref{tab:MSFE}. The results indicate that with a foreground threshold of 0.1, simultaneous use of ${\bold{F}^{4}}$ and ${\bold{F}^{8}}$ achieves the best enhancement effects.

\begin{table}[t]
\centering
\caption{Comparison with cross-modal distillation methods. The improvements caused by distillation are shown in \textcolor[RGB]{255,0,0}{red}.}
\label{tab:cm1}
\renewcommand\arraystretch{1.3}
\scriptsize
\setlength{\tabcolsep}{2mm}{
\begin{tabular}{l|c|c|c|cc}
\hline
Methods & Backbone & Teacher Modality & Student  & mAP & NDS \\ 
\hline
BEVDistill & ResNet50 & LiDAR & BEVDepth  & 0.330(\textcolor[RGB]{255,0,0}{+1.3\%}) & 0.452(\textcolor[RGB]{255,0,0}{+1.2\%})  \\
UniDistill & ResNet50 & LiDAR+Camera & BEVDet  & 0.296(\textcolor[RGB]{255,0,0}{+3.2\%}) & 0.393(\textcolor[RGB]{255,0,0}{+3.2\%})  \\
\rowcolor{gray!20}
FSD-BEV & ResNet50 & LiDAR+Camera & BEVDepth & 0.403(\textcolor[RGB]{255,0,0}{+4.0\%}) & 0.526(\textcolor[RGB]{255,0,0}{+4.0\%}) \\
\hline
TiG-BEV & ResNet101 & LiDAR & BEVDepth & 0.440(\textcolor[RGB]{255,0,0}{+2.4\%}) & 0.544(\textcolor[RGB]{255,0,0}{+2.3\%}) \\
X${}^3$KD & ResNet101 & LiDAR & BEVDepth & 0.448(\textcolor[RGB]{255,0,0}{+3.9\%}) & 0.553(\textcolor[RGB]{255,0,0}{+2.2\%}) \\
DistillBEV & ResNet101 & LiDAR+Camera & BEVDepth & 0.450(\textcolor[RGB]{255,0,0}{+3.8\%}) & 0.547(\textcolor[RGB]{255,0,0}{+1.2\%}) \\
\rowcolor{gray!20}
FSD-BEV & ResNet101 & LiDAR+Camera & BEVDepth & 0.488(\textcolor[RGB]{255,0,0}{+7.6\%}) & 0.589(\textcolor[RGB]{255,0,0}{+5.4\%}) \\
\hline
\end{tabular}}
\end{table}

\begin{table}[t]
\centering
\caption{Comparison of distillation performance for different model magnitudes.} 
\label{tab:cm2}
\renewcommand\arraystretch{1.3}
\scriptsize
\setlength{\tabcolsep}{3mm}{
\begin{tabular}{l|c|c|c|c|c}
\hline
\multirow{2}*{Methods} & \multirow{2}*{Backbone} & \multicolumn{2}{c|}{Teacher} & \multicolumn{2}{c}{Student}  \\ 
\cline{3-6}
 & & mAP & NDS & mAP & NDS\\
\hline
\multirow{2}*{DistillBEV\cite{wang2023distillbev}} & ResNet50  & 0.671 & 0.780 & 0.403 & 0.510\\
\cline{2-6}
& ResNet101 & 0.671 & 0.780 & 0.450$_{+4.7\%}$ & 0.547$_{+3.7\%}$\\ 
\hline
\multirow{2}*{Ours}  & ResNet50 & 0.584 & 0.641 & 0.403 & 0.526 \\
\cline{2-6}
  & ResNet101 & 0.680 & 0.717 & 0.488$_{+8.5\%}$ & 0.589$_{+6.3\%}$\\ 
\hline
\end{tabular}}
\end{table}

\subsection{Comparison with Cross-Modal Distillation}

We compare FSD-BEV with methods that apply cross-modal distillation to improve the camera-only BEV-based 3D detectors. From the results in \cref{tab:cm1}, it can be found that the simple foreground self-distillation framework outperforms other distillation methods that adopt complicated strategies. The advantage of FSD-BEV is more significant when the ResNet101 is chosen as the image backbone and a total of 7.6\% and 5.4\% in mAP and NDS is achieved, which largely exceeds the distillation gains of other methods.

The joint progress of the student model and the teacher model is a benefit characteristic of FSD-BEV. To evaluate the profit of this characteristic, we chose the state-of-the-art cross-modal distillation approach, DistillBEV \cite{wang2023distillbev}, as the control group. Each model is configured with two sets of control experiments, the model in one set using ResNet50 as the backbone network and the models in the other set using ResNet101. The comparison results are shown in \cref{tab:cm2}.

Under the setting of ResNet50, the student accuracy of FSD-BEV is roughly comparable to that of DistillBEV. However, when we switch to a larger backbone, the student accuracy of DistillBEV improves by 4.7\% and 3.7\% in mAP and NDS, FSD-BEV improves by 8.5\% and 6.3\%. It is worth noting that our teacher does not exhibit higher performance compared to the teacher of DistillBEV, indicating our method's higher distillation efficiency. On the other hand, compared to cross-modal methods with a fixed teacher accuracy, our teacher accuracy increased by 9.4\% in mAP and 7.6\% in NDS when a larger backbone network is employed. This growth aligns with the magnitude of improvement seen in the student. Such cooperative growth enables our approach to maintain stable distillation performance.

\section{Conclusion}

In order to narrow the performance gap between LiDAR-based 3D detectors and BEV-based multi-view 3D detectors, several methods utilize cross-modal knowledge distillation attempts to transfer beneficial information from the extra teacher model to the student model. However, such approaches always suffer from feature distribution discrepancies originating from different data modalities and network structures. In this paper, we propose a Foreground Self-Distillation (FSD) scheme that effectively alleviates this issue without the need for pre-trained teacher models or cumbersome distillation strategies. Additionally, we design two Point Cloud Intensification (PCI) strategies to compensate for the sparsity of point clouds by frame combination and pseudo point assignment, further intensifying the performance of the teacher model. Finally, we develop a Multi-Scale Foreground Enhancement (MSFE) module to extract and fuse multi-scale foreground features by predicted elliptical Gaussian heatmap, further improving the model's performance.

\section*{Acknowledgement}

This work was supported by Zhejiang Provincial Natural Science Foundation of China (No. LD24F020016), "Pioneer" and "Leading Goose" R\&D Program of Zhejiang (No. 2024C01020), the National Natural Science Foundation of China (No. 62176017, No. 62302031), and ZEEKR Intelligent Technology Co., Ltd.

%
%
\bibliographystyle{splncs04}
\bibliography{main}

\newpage
\appendix
\renewcommand\thetable{\Alph{table}}
\renewcommand\thefigure{\Alph{figure}}
\setcounter{figure}{0}
\section*{Supplementary Material}

This supplementary material provides more implementation details on FSD-BEV in Sec.~\ref{sec:A}, more experiment results in Sec.~\ref{sec:B} and visualization results in Sec.~\ref{sec:C}.

\section{More Implementation Details}
\label{sec:A}
\subsection{Data Augmentation}
We first perform random scaling on the input images with a scaling factor in the range of $\left [ 0.5, \; 1.25 \right ]$. Then, we crop the images according to the input size, followed by flipping operations with a probability of 0.5. Finally, we rotate the images within the range of $\left [ -5.4^{\circ}, \; 5.4^{\circ} \right ]$ to obtain the augmented input images. Similar to BEVDepth~\cite{li2023bevdepth}, we also perform data augmentation on BEV features. The rotation range is $\left [ -22.5^{\circ}, \; 22.5^{\circ} \right ]$, the scaling factor ranges from $\left [ 0.95, \; 1.05 \right ]$, and flipping is applied independently along the X and Y axes with a probability of 0.5.

\subsection{Details of Training}
During training, we generate ground truth heatmaps by drawing elliptical Gaussian distributions on the original image size and then performing rigid body transformations similar to those applied to the image. We compute the depth loss using Cross Entropy Loss. The center head in Centerpoint~\cite{yin2021center} is employed as the detection head, using Gaussian Focal Loss to supervise the heatmap of BEV features and L$_{1}$ Loss as the regression loss.

We set the detection region along the X and Y axes to $\left [ -51.2, \; 51.2 \right ]$ and along the Z axis to $\left [ -5, \; 3 \right ]$. When our image input size is 256 $\times$ 704, the BEV features are divided into sizes of 128 $\times$ 128. However, when we use larger image input sizes, the BEV size is increased to 256 $\times$ 256.

\begin{figure}
  \centering
    \includegraphics[width=0.9\linewidth]{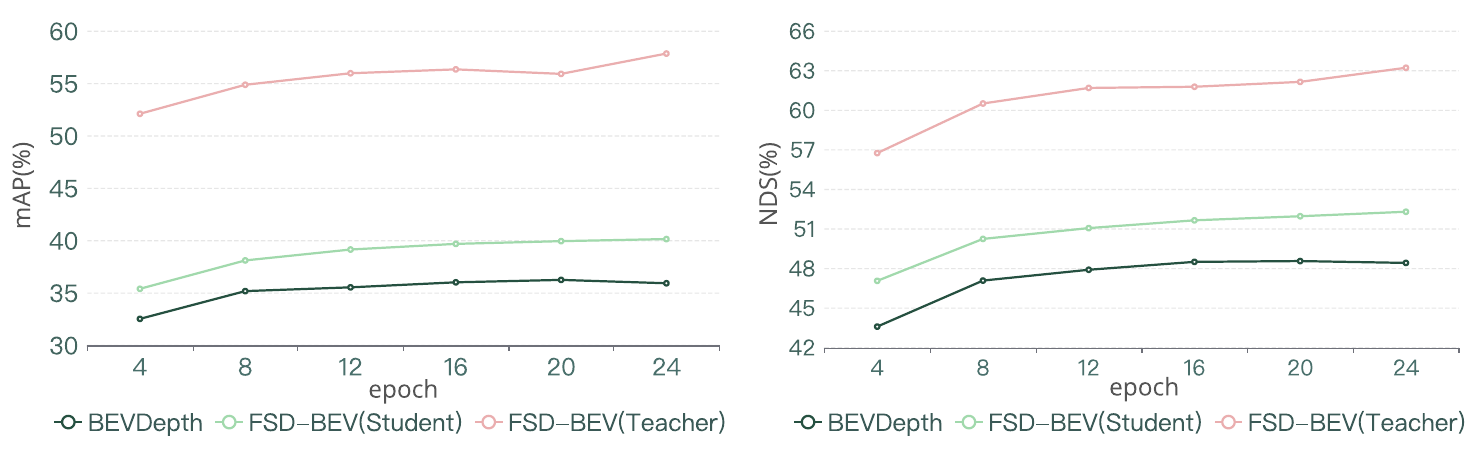}
  \caption{Comparison of performance between baseline (BEVDepth) and FSD-BEV during training. FSD-BEV is divided into student and teacher branches, and we evaluate mAP and NDS on the nuScenes $val$ set.}
  \label{Fig.mapnds}
\end{figure}

\section{More Experiment Results}
\label{sec:B}

\subsection{Performance Analysis of the Training Process}
We conduct a comparative analysis of the performance between the baseline (BEVDepth) and FSD-BEV over the entire training process. Both of them use ResNet50 as the backbone network and are trained for 24 epochs with the CBGS strategy, and their performance is depicted in \cref{Fig.mapnds}. During the training process, it is observed that the precision of the FSD-BEV's teacher branch increases and consistently provides high-quality guidance to the student branch, resulting in significant improvement compared to the baseline. After training for 20 epochs, there is a slight decrease in the precision of BEVDepth, indicating the occurrence of overfitting. On the contrary, FSD-BEV continues to demonstrate a growth trend, which is attributed to the accurate depth information provided by the teacher branch.

\begin{figure}
  \centering
    \includegraphics[width=0.8\linewidth]{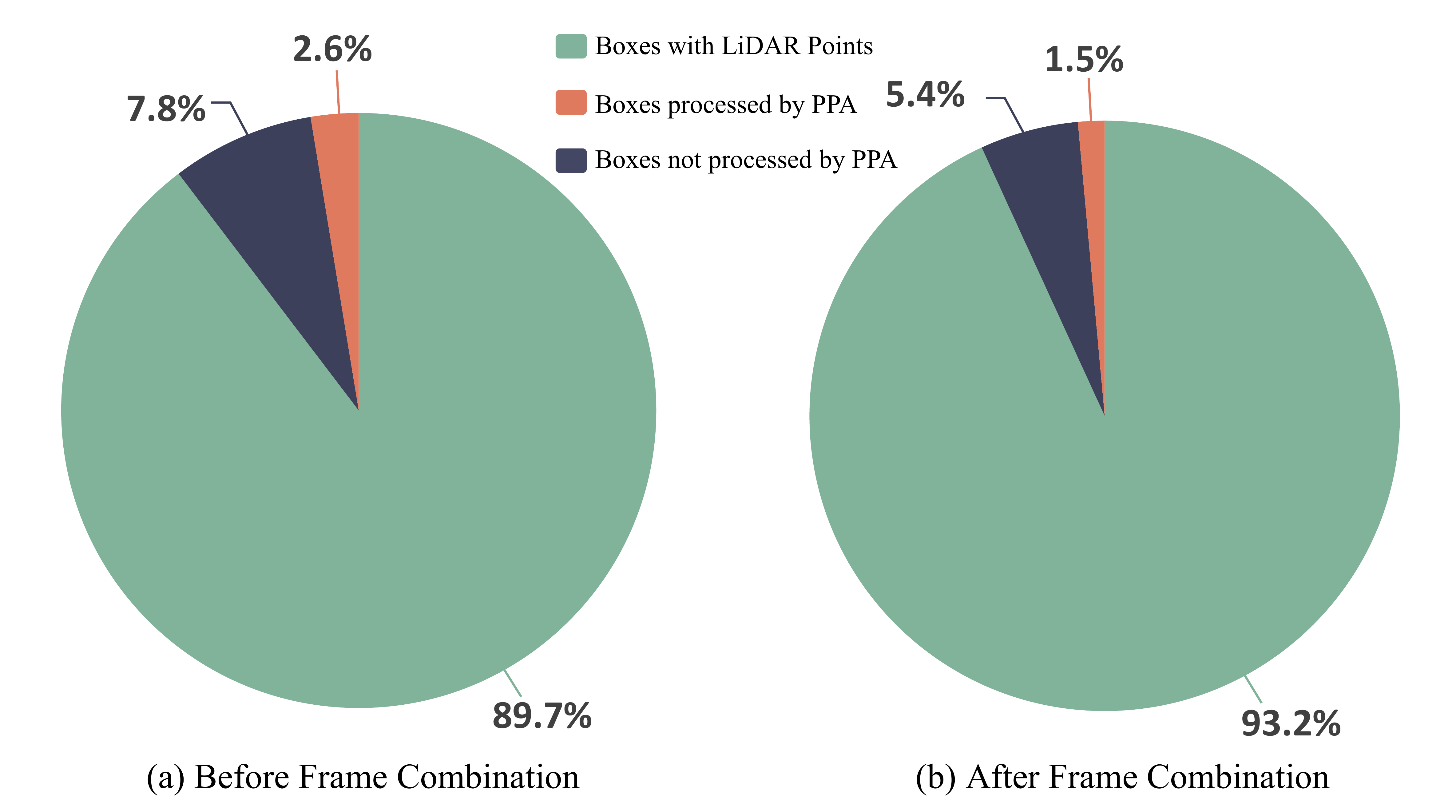}
  \caption{Statistics of the GT boxes modified by Point Cloud Intensification. (a) illustrates the proportion of benefited GT boxes solely after applying PPA, while (b) shows the proportion of benefited GT boxes under the combined action of FC and PPA.}
  \label{Fig.pci}
\end{figure}
\subsection{Statistics of Point Cloud Intensification}
We quantify the efficacy of Point Cloud Intensification (PCI) by counting the number of intensified ground truth (GT) 3D boxes in the nuScenes $train$ dataset. The proportions of the benefited GT boxes are illustrated in \cref{Fig.pci}. It can be observed that 10.4\% of the GT boxes do not carry LiDAR points, which is detrimental to generating high-quality teacher BEV. After applying PPA, 2.6\% of the GT boxes are appropriate for supplementing pseudo points, mitigating the loss of objects information to a certain extent. If the Frame Combination (FC) is first applied, the proportion of GT boxes without LiDAR points drops to 6.9\%, and the proportion of the GT boxes benefit from PPA drops to 1.5\%, demonstrating that FC can perform some of the functions of PPA. However, there are still 5.4\% of the GT boxes can not be supplemented by PCI. They could have bad visibility or be located too distant, and boldly intensifying them may introduce inaccurate information.

\begin{figure}
  \centering
    \includegraphics[width=1\linewidth]{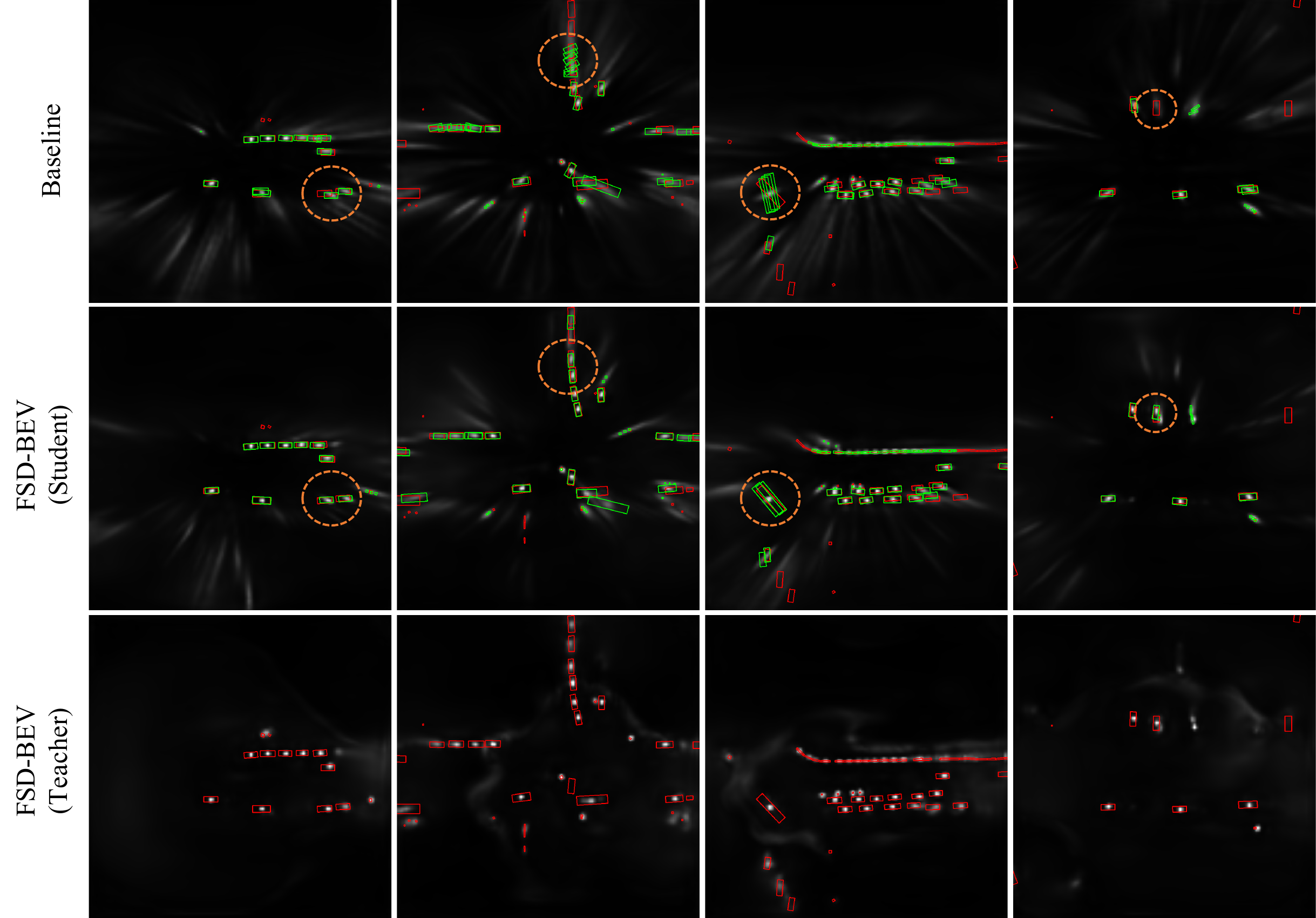}
  \caption{Visualization results of FSD-BEV and baseline (BEVDepth) on BEV heatmaps. The red and green boxes represent the ground truth and predicted results, while orange circles denote improvement examples of FSD-BEV compared to the baseline.}
  \label{Fig.vishm}
\end{figure}
\section{Visualization}
\label{sec:C}
As shown in \cref{Fig.vishm}, we visualize the predicted BEV heatmaps and bounding boxes of different models. The BEV heatmaps predicted by the teacher branch of FSD-BEV match the GT boxes well, reflecting the effectiveness of hard labels. Compared with the baseline, the heatmaps predicted by the student branch of FSD-BEV are closer to the teacher's high-quality BEV heatmaps, which leads to more precise predicted boxes. Since the heatmaps are obtained from encoded $\hat{\bold{B}}_{s}$ and $\hat{\bold{B}}_{t}$ mentioned in the main text, it indicates that our distillation scheme works well on forcing $\hat{\bold{B}}_{s}$ to imitate $\hat{\bold{B}}_{t}$.


\end{document}